\documentclass[lettersize,journal]{IEEEtran}
\usepackage{amsmath,amsfonts}
\usepackage{algorithmic}
\usepackage{algorithm}
\usepackage{array}
\usepackage[caption=false,font=normalsize,labelfont=sf,textfont=sf]{subfig}
\usepackage{textcomp}
\usepackage{stfloats}
\usepackage{url}
\usepackage{verbatim}
\usepackage{graphicx}
\usepackage{cite}
\usepackage{rotating}
\usepackage[normalem]{ulem}

\usepackage{url}
\usepackage{epsfig}
\usepackage{tabularx}
\usepackage{xcolor}
\begin{document}

\title{A Spatiotemporal Extension of the Neuromorphic DBSCAN  Implementation}

\author{Charles P. Rizzo and James S. Plank
\begin{center}
Department of Electrical Engineering and Computer Science \\
University of Tennessee \\
Knoxville, TN 37996 \\
\mbox{} \\
Corresponding authors: Charles P. Rizzo {\tt crizzo@utk.edu} or James S. Plank: {\tt jplank@utk.edu}
\end{center}
}

\maketitle

\begin{abstract}
DBSCAN is an algorithm that denoises and clusters data. In prior work, we implemented the DBSCAN algorithm neuromorphically, introducing two constructions termed ``flat'' and ``systolic''. The ``flat'' construction prioritizes throughput, while the ``systolic'' construction trades time for space resulting in a smaller, more hardware-friendly architecture at the cost of throughput.
In this work, we offer spatiotemporal extensions of these two constructions to better leverage the spatiotemporal nature of event sensor data. Moreover, as in our prior work, we discuss partial or segmented implementations that further leverage time for space when hardware resources are constrained. All network constructions are provided as open-source
implementations. 

\end{abstract}


\section{Introduction}

\noindent The Density-based Spatial Clustering of Applications (DBSCAN) algorithm is a clustering and denoising algorithm that was popularly introduced by Esker, Kriegel, Sander, and Xu~\cite{eks:96:dbs}. The algorithm uses two parameters -- $\epsilon$, a radial distance, and $minPoints$, a measure of density -- to determine whether each data point in a set of correlated data should be classified as a \textbf{core} point of a cluster, a \textbf{border} point of a cluster, or a \textbf{noise} point that can be discarded. The result of running the algorithm on a set of data is a set of non-rigidly bound clusters with noisy outliers removed, depending on the values of $\epsilon$ and $minPoints$.

Our primary interest with the DBSCAN algorithm is to apply it to event-based camera data while simultaneously leveraging the low Size, Weight, and Power (SWaP) of neuromorphic computing. Neuromorphic computing uses spiking neurons and synapses to propagate information through a neural network temporally, drawing heavy inspiration from the human brain and its unparalleled efficiency. As an event-based processing paradigm, it is especially appealing for deployment at the edge, and its pairing with event-based sensors, such as event cameras, is a natural use case. We initially explored a neuromorphic implementation of the DBSCAN algorithm in~\cite{r:24:eeb,rsp:24:sbf}. This work culminated into a formal definition of DBSCAN as a neuromorphic implementation in~\cite{rp:24:ani}. In~\cite{rp:24:ani}, we presented two different constructions of the DBSCAN algorithm neuromorphically, termed ``flat'' and ``systolic'', where the former prioritizes processing speed and the latter hardware resources. We characterized several networks in terms of runtime and size for various $\epsilon$ and $minPoints$ values, explored partial implementations of each construction to further reduce hardware resource usage, and open-sourced the software that produces these spiking neural networks:~\url{https://github.com/TENNLab-UTK/dbscan}.

Others have applied clustering algorithms like DBSCAN  directly to event camera data~\cite{nlr:23:dotie,hkl:22:ecdt}, and there has also been work specifically in clustering events as they're asynchronously generated by the camera. This is done by Rodriguez {\em et al}~\cite{rem:20:asynchronous}, which features UAV-mounted event-based cameras to detect intruders and by Helwa, Kratchman, and Bishop, where they present EventStream, a data stream clustering algorithm (like eCDT~\cite{hkl:22:ecdt}), where DBSCAN plays a crucial role in clustering the final event clusters~\cite{hkb:24:eventstream}. Schraml and Belbachir also present a clustering algorithm derived from DBSCAN~\cite{sb:10:spatio}.

However, perhaps most related to our goal of temporally extending the DBSCAN algorithm neuromorphically is work performed by Zhang {\em et al}. In~\cite{zgs:23:neuromorphic}, the authors use DBSCAN spatiotemporally where the distance formula for event locality includes the temporal domain. They demonstrated the effectiveness of their 3D-augmented DBSCAN algorithm for denoising by injecting various quantities of noise into different datasets (N-Cars~\cite{sbb:18:hats}, ASL-DVS~\cite{bca:19:asldvs}, N-MNIST~\cite{ojc:15:nmnist}, POKER-DVS~\cite{sl:15:pokerdvs}) and applying their algorithm (among others for comparison~\cite{d:08:frame,lbl:15:design,kk:18:n,gd:22:latency,ffl:20:event,wml:20:probabilistic}) to the noise-enhanced datasets. They then compared the top-1 accuracy of a trained ResNet-18 model~\cite{hzr:16:resnet} on the denoised versions of the datasets, as produced by the different algorithms, to evaluate each algorithm's denoising effectiveness in terms of trained model accuracy. They showed that as the applied noise percentage increased, their 3D-augmented DBSCAN algorithm was the most effective at producing the better performing classifier. This work in particular serves as motivation to extend our current neuromorphic definition of the DBSCAN algorithm~\cite{rp:24:ani} so that it may operate spatiotemporally, thereby reaping the benefits of low-SWaP neuromorphic hardware and spatiotemporal event correlation.

In this work, we extend our prior work by defining variants of the two aforementioned constructions that can operate on spatiotemporal data or streams of events over time from an event camera. The spatiotemporal constructions, like the initial implementations, are precisely defined and characterized so that they may be extended to other generic neuromorphic computing platforms and realized neuromorphic hardware like Loihi2~\cite{ofr:21:loihi2}. Practical considerations are also discussed detailing network runtimes, parameters, and scalability. The authors provide the code used for this work as open-source in order to encourage adoption in the neuromorphic community as a benchmark for operating on event sensor data in real-time on embedded hardware.

\section{Background}

\subsection{DBSCAN Algorithm}
\noindent Our description of the DBSCAN algorithm is identical to what is described in~\cite{rp:24:ani}. The conventional, 2D implementation of the algorithm operates spatially on some~$R\times{C}$ input. Because our application of interest is event-based cameras, our input is composed of 0s and 1s where events of both positive and negative polarity are assigned a value of 1 and all other input locations for some frame of input are assigned a value of 0 and are ignored.

The algorithm has two parameters: a radial parameter $\epsilon$ and a neighborhood density measure $minPoints$.  The DBSCAN algorithm classifies every event in an input as one of {\em Core}, {\em
Border} and {\em Noise} according to its two parameter values.

We reiterate the following assumptions and classification definitions exactly as they appear in~\cite{rp:24:ani}: 

\begin{itemize}
\item Let $E_{r,c}$ be the event at row~$r$ and column~$c$ on the grid.
\item The neighborhood~$N$ of $E_{r,c}$ is composed of all events~$E_{i,j}$ where
      $|i-r| \le \epsilon$ and $|j-c| \le \epsilon$.
\item Let~$N$ be the maximum size of a neighborhood of an event:
      $N = (2\epsilon+1)^2$.  $N$ is the maximum size, because neighborhoods
      are smaller at the borders of the grids, where, for example, an event on the top-left
      corner of the grid has no neighborhood above or left of it. $N$ is derived from the Chebyshev distance algorithm, which can be thought of as hop distance in all directions.
\item $minPoints$ is a number between 1 and~$N$, and is a measure of the density of clustering.
\end{itemize}

There are three event classifications, and they are determined as follows:

\begin{enumerate}
\item {\em Core}: An event is classified as {\em Core} if there are at least $minPoints$ events in its neighborhood.
\item {\em Border}: An event is classified as {\em Border} if it is not {\em Core}, but there is at least
      one {\em Core} event in its neighborhood.
\item {\em Noise}: An event is classified as {\em Noise} if it is neither {\em Core} nor {\em Border}.
\end{enumerate}

\subsection{Neuromorphic Processor}\label{sec:risp}
\noindent We also use the same spiking neural network definition presented in~\cite{rp:24:ani}: the RISP neuroprocessor model~\cite{pzg:22:risp,pdg:24:risp}. It features a variety of the leaky integrate-and-fire neuron (LIF) where the leak values  are set either to ``all'' or ``none'' for every network cycle. A leak value of ``all'' means that a neuron leaks away all of its accumulated activation potential if it does not fire at the end of each network timestep. A leak value of ``none'' implies the opposite: a neuron leaks no charge at the end of every timestep. 

Every neuron accumulates charge up until its activation potential meets its threshold value. When a neuron's activation potential meets or exceeds its threshold at the end of a network timestep or intgration cycle, that neuron fires onto its outgoing synapses and its activation potential is reset to whatever its minimum potential value is defined as -- typically zero. All neurons have programmable threshold, leak, and minimum potential values 

The synapses are composed of two properties: weight and delay. When a neuron fires onto one of its outgoing synapses, that unit of charge will carry a value equal to the synapse's weight and will arrive at the post-synaptic neuron exactly \textit{delay} timesteps later. For this work, as in~\cite{rp:24:ani}, all neuronal and synaptic parameters are integers. Additionally, all of the networks presented in this work can be deployed to hardware such as microcontrollers~\cite{gmp:25:gsn} or FPGAs\cite{pdg:24:risp} using software defined as part of the TENNLab software suite~\cite{psb:18:ten,psr:24:fo}.

\subsection{The ``flat'' construction}\label{sec:flat}
\noindent The flat implementation of the DBSCAN algorithm is composed of $5 \times R \times C$ neurons. Every neuron is denoted as belonging to a collection or layer and is parameterized by~$r$ and~$c$, the row and column indices of the neuron. Some sample input of events to the SNN is denoted as~$E$. In terms of spike propagation, this construction is a strictly feed-forward architecture that can be thought of as layered and having skip connections where each layer contains $RC$ neurons. Table~\ref{tab:dbscan_neurons} provides a general description of the five collections of neurons, their threshold values, and the notations that will be used to refer to them as throughout the remainder of this work. The reader is referred to~\cite{rp:24:ani} for a more detailed explanation about the neuron collections' connectivity and fire times.

\begin{table*}[]
\caption{Neuron Collections}
\label{tab:dbscan_neurons}
\begin{tabular}{l|cccl}
\textbf{}            & \textbf{Notation} & \textbf{Type} & \textbf{Threshold} & \multicolumn{1}{c}{\textbf{Description}}                                                                                                                                                                                                     \\ \hline
\textbf{Input}       & $Input_{r,c}$     & Input         & 1                  & \begin{tabular}[c]{@{}l@{}}Input event $E_{r,c}$ is applied to $Input_{r,c}$.\end{tabular}                                                                                                                                \\ \\
\textbf{Counts} & $C_{r,c}$        & Hidden        & $minPoints - 1$    & \begin{tabular}[c]{@{}l@{}}Counts number of events in event $E_{r,c}$'s neighborhood \textit{N} (excluding event $E_{r,c}$). \end{tabular} \\ \\
\textbf{Core}        & $Core_{r,c}$      & Output        & 2                  & \begin{tabular}[c]{@{}l@{}}Both $Input_{r,c}$ and $C_{r,c}$ must fire for $Core_{r,c}$ to fire.\end{tabular}                                                                                \\ \\
\textbf{Border-Cores}  & $B_{r,c}$        & Hidden        & 1                  & \begin{tabular}[c]{@{}l@{}}Number of Core events in event $E_{r,c}$ 's neighborhood \textit{N} (excluding event $E_{r,c}$).\end{tabular}              \\ \\
\textbf{Border}      & $Border_{r,c}$    & Output        & 2                  & \begin{tabular}[c]{@{}l@{}}Both $Input_{r,c}$ and $B_{r,c}$ must fire for $Border_{r,c}$ to fire. $Core_{r,c}$ must \textbf{not} fire.\end{tabular}                            
\end{tabular}
\end{table*}

\begin{figure}[ht]
\begin{center}
\includegraphics[width=\linewidth,trim={7.5cm 7cm 7.5cm 7cm},clip]{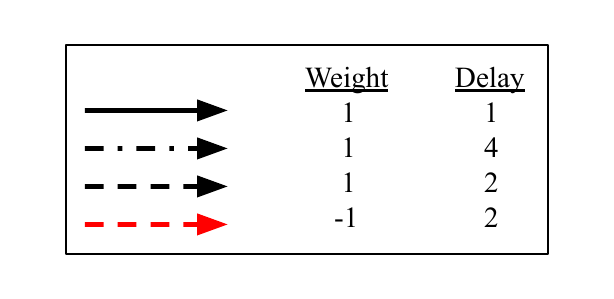}
\caption{Legend for parsing synaptic connectivity between neuron collections in Figures~\ref{fig:flat_connectivity},~\ref{fig:systolic_connectivity}, and~\ref{fig:temporal_flat_connectivity}.}
\label{fig:legend}
\end{center}
\end{figure}

\begin{figure*}[ht]
\begin{center}
\includegraphics[width=\textwidth,trim={1.7cm 7cm 1.7cm 6cm},clip]{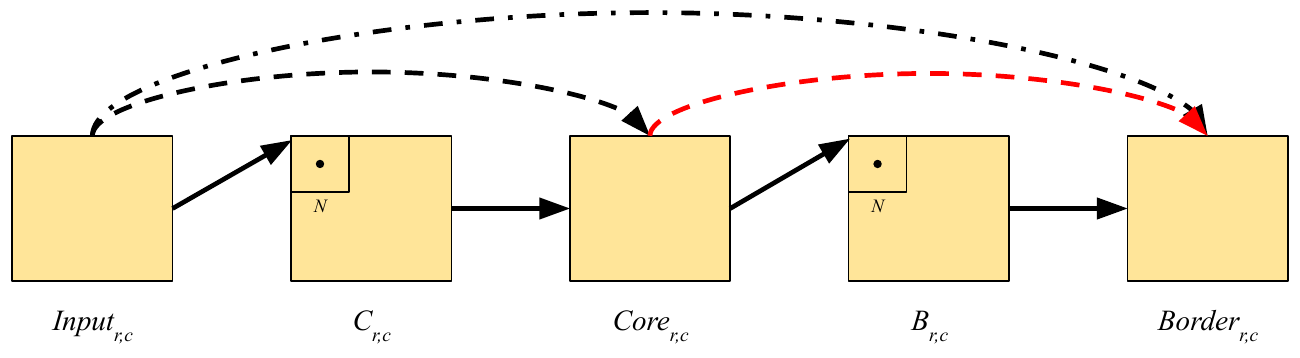}
\caption{Illustration showing the connectivity between the five neuron collections for the flat algorithm.}
\label{fig:flat_connectivity}
\end{center}
\end{figure*}

Functionally speaking, the \textbf{Input} ($Input_{r,c}$) neurons receive the input events ($E$) from some sensor (such as an event-based camera) and fire when an event is received. The \textbf{Count} ($C_{r,c}$) neurons receive spikes from the $Input_{r,c}$ neurons and check if there are at least $(minPoints - 1)$ other spikes (or events from the event camera) in the neighborhoods $N$ of each $E_{r,c}$ event. If for some event $E_{r,c}$, there exists $(minPoints - 1)$ other events in $N_{r,c}$, neuron $C_{r,c}$ will fire such that $Core_{r,c}$ (also receiving the input event $E_{r,c}$ from $Input_{r,c}$) will fire, classifying event $E_{r,c}$ as a \textbf{core} event. As with the \textbf{Count} ($C_{r,c}$) neurons, the \textbf{Border-Cores} ($B_{r,c}$) neurons will receive spikes from the $Core_{r,c}$ neurons and check if there is at least 1 \textbf{core} event in the neighborhoods $N$ of each $E_{r,c}$ event. If so, the $B_{r,c}$ neurons will fire and send spikes to the $Border_{r,c}$ neurons which will fire if and only if an event $E_{r,c}$ arrives from $Input_{r,c}$, the corresponding $B_{r,c}$ neuron has fired -- indicating a core event classification in the neighborhood $N_{r,c}$ has been made --, and the corresponding $Core_{r,c}$ neuron has {\em not} fired. The $Core_{r,c}$ neuron sends inhibitory charge to the corresponding $Border_{r,c}$ neuron when it fires to prevent a \textbf{core} event $E_{r,c}$ from being misclassified as a \textbf{border} event. Figure~\ref{fig:flat_connectivity} illustrates the synaptic inter-connectivity of the different collections at a high level, and Figure~\ref{fig:legend} depicts a legend that details how to parse the synaptic weights and delays of the connectivity.

This architecture is illustrated in greater detail with an example in~\cite{rp:24:ani} detailing the neuron collections' activation potentials during the timesteps that they fire in addition to formalizing inter-collection or inter-layer synaptic connectivity.

From~\cite{rp:24:ani}, we know that the flat algorithm produces spiking neural networks that may not fit on available hardware. At $5RC$ neurons and roughly~$(3+2N)RC$ synapses, even though these networks can process an input frame in as little as five network timesteps and perfect pipelining is supported, an input frame whose dimensions are the size of Inivation's Davis346 camera~\cite{i:19:346} ($260 \times 346)$ would result in a spiking neural network with nearly 0.5 million neurons and nearly 15 million synapses. Newer event cameras with HD resolutions would greatly exceed these neuronal and synaptic totals.

\subsection{The ``systolic'' construction} \label{sec:systolic}
\noindent With the systolic implementation, instead of collections that are $R \times C$ in size, our collections are of size $R$. Two of the collections, the \textbf{Input} and \textbf{Core} neurons, are also parameterized by $\epsilon$. Instead of applying and processing one $R \times C$ region of input in one timestep, we apply the input in a streaming, timeseries manner such that $R$ inputs are applied over $C$ timesteps. This results in fewer inferences per second compared to our flat implementation, but it also produces a much smaller spiking neural network. With fewer neurons overall, the systolic algorithm is optimal in terms of resource utilization. The collections of neurons have the same names and functions as in the flat specification, so Table~\ref{tab:dbscan_neurons} still applies. Generally speaking, the systolic implementation reduces a 2-dimensional input to a 1-dimensional input where information along the second dimension is applied to the 1D vector of input neurons temporally. Recalling from~\cite{rp:24:ani}, the neuron collections are parameterized by:

\begin{eqnarray*}
r & | & 0 \le r < R \\
e & | & -\epsilon \le e \le \epsilon \\
\end{eqnarray*}

\begin{figure*}[ht]
\begin{center}
\includegraphics[width=\textwidth,trim={1.5cm 5cm 3cm 5cm},clip]{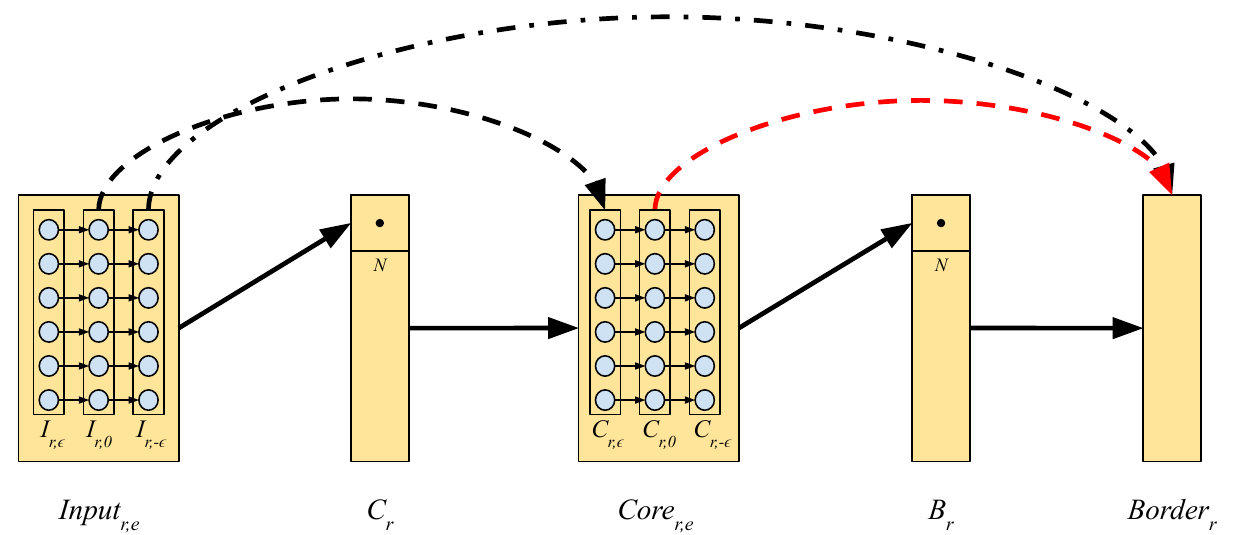}
\caption{Illustration showing the connectivity between the five neuron collections for the systolic algorithm.} 
\label{fig:systolic_connectivity}
\end{center}
\end{figure*}

Figure~\ref{fig:systolic_connectivity} illustrates the high level synaptic connectivity for the systolic algorithm as is done in Figure~\ref{fig:flat_connectivity} for the flat algorithm.

To classify an event $E_{r,c}$, its entire neighborhood of event activity~$N$ must be present in the network during the same timestep. With the flat algorithm, the neighborhoods~$N$ for all events $E$ would be applied to the $Input$ neurons during the first timestep. With the systolic algorithm, we apply one column's worth of values to the network per timestep, so we need a total of $R(2\epsilon + 1)$ of $Input$ neurons to hold in network memory the neighborhoods~$N$ of all events in column $c$. The only neurons to which external input can be applied in the $Input_{r,e}$ collection are the $Input_{r,\epsilon}$ neurons. The $Core_{r,e}$ neurons operate similarly to the $Input_{r,e}$ on incoming spike data from the $Input_{r,0}$ and $C_{r}$ neurons. The $C_{r}$, $B_{r}$, and $Border_{r}$ neurons are all simply one dimensional vectors of $R$ neurons, whereas the $Input$ and $Core$ neuron collections are composed of $R(2\epsilon + 1)$ neurons each. The reader is again referred to~\cite{rp:24:ani} for a more detailed walkthrough of the neuron collections' connectivity and execution through time on a specific example.

Systolic networks contain $R(4\epsilon + 5)$ neurons and roughly~$R(2N + 4\epsilon + 3)$ synapses~\cite{rp:24:ani}. Using Inivation's Davis346 event-based camera again as an example, a $260\times{346}$ resolution input, with $\epsilon = 4$, yields a network with fewer than six thousand neurons and fifty thousand synapses. Compared to the flat algorithm, these neuron and synaptic totals are much more tractable on available neuromorphic hardware. However, the time-to-solution on one input frame is now $C + 2\epsilon + 4 = 346 + 2(4) + 4 = 358$ timesteps compared to the flat algorithm's $5$ timesteps. As with the flat algorithm, the systolic implementation pipelines well, but not perfectly. $2\epsilon$ timesteps must be added to every DBSCAN calculation in order to clear the remaining charge in the network from the last processed column and its neighborhood.  In other words, the same systolic DBSCAN network can operate on a new input every~$C+2\epsilon$ timesteps.

\section{Spatiotemporal DBSCAN}\label{sec:3d_dbscan}

\noindent DBSCAN is strictly defined as a spatial clustering algorithm; however, event-based sensors are inherently temporal sensors that excel at capturing data at high speeds over time. A popular method for processing event-based data is to create event frames and operate on the accumulation of events as a 2D frame. As the aforementioned DBSCAN algorithms require a $R \times C$ grid of inputs, creating event frames as inputs and spatially clustering and denoising the sequential, independent event frames is a sensible workflow. However, for event-based cameras, captured motion and the events generated by it are inherently spatiotemporally related. This means that events that belong to some moving object captured in the current frame at time $t$ are likely related to nearby events that occurred in frame $t-1$.  In~\cite{zgs:23:neuromorphic}, Zhang et al. demonstrate the efficacy of a 3D-augmented DBSCAN-derived algorithm to denoise event camera data. They show both qualitatively and quantitatively that their algorithm performs competitively compared to, and in many cases better than, other popular event camera denoising algorithms~\cite{d:08:frame,lbl:15:design,kk:18:n,ffl:20:event,wml:20:probabilistic}.

Inspired by this extension of the DBSCAN algorithm, we also implement temporal extensions of both the flat and systolic DBSCAN constructions presented in sections~\ref{sec:flat} and~\ref{sec:systolic}. Zhang et al. present a generalized higher-dimensional extension of the DBSCAN algorithm where the event being classified is some central point in a redefined neighborhood~$N$, which is a 3D voxel of events that is spatially parameterized by radius $\epsilon_{s}$ and temporally parameterized by radius $\epsilon_{t}$. Instead of looking only at the two-dimensional square of neighboring spatial event activity, their implementation looks at the three-dimensional rectangular prism of spatiotemporal neighboring event activity. While this implementation is reasonable, it is not ideal for deployment in a live environment where minimal system latency is desired. The implementation of Zhang et al. suffers, at minimum, a classification latency of $\epsilon_{t}$. Classifying some event $E_{r,c}$ that occurs at time $t$ requires the event voxel neighborhood~$N$ that spans from time $t - \epsilon_{t}$ to time $t + \epsilon_{t}$.

\begin{figure}[ht]
\begin{center}
\includegraphics[trim={8.5cm 4.5cm 4.5cm 5.5cm},clip]{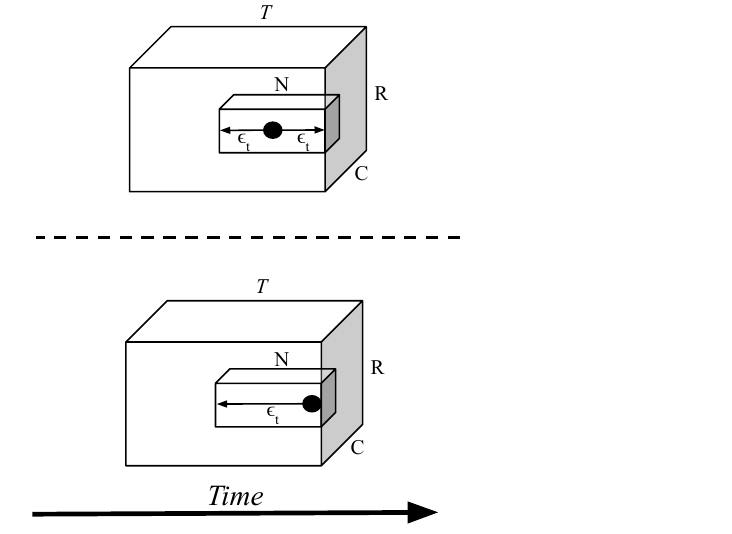}
\caption{(Top) Spatiotemporal DBSCAN-based algorithm presented by Zhang et al.~\cite{zgs:23:neuromorphic}. (Bottom) Spatiotemporal DBSCAN-based algorithm presented in this work.}
\label{fig:3d_dbscan_diagram}
\end{center}
\end{figure}

For our implementation, instead of treating $\epsilon_{t}$ as a radius, we treat it as a temporal look-back. With this implementation, all events occurring at time $t$ are classified immediately without dependence on event activity that has not yet occurred. Figure~\ref{fig:3d_dbscan_diagram} illustrates the difference between the algorithm derived by Zhang et al. in~\cite{zgs:23:neuromorphic} (top) and the algorithm presented in this work (bottom). We use $e_{t}$ to denote events at a particular time or frame as $E_{e_{t},r,c}$. We define $e_{t}$ as follows:

\begin{eqnarray*}
e_{t} & | & 0 \le e_{t} < \epsilon_{t} \\
\end{eqnarray*}

\begin{figure*}[ht]
\begin{center}
\includegraphics[width=\textwidth,trim={1.0cm 6.5cm 1.5cm 2cm},clip]{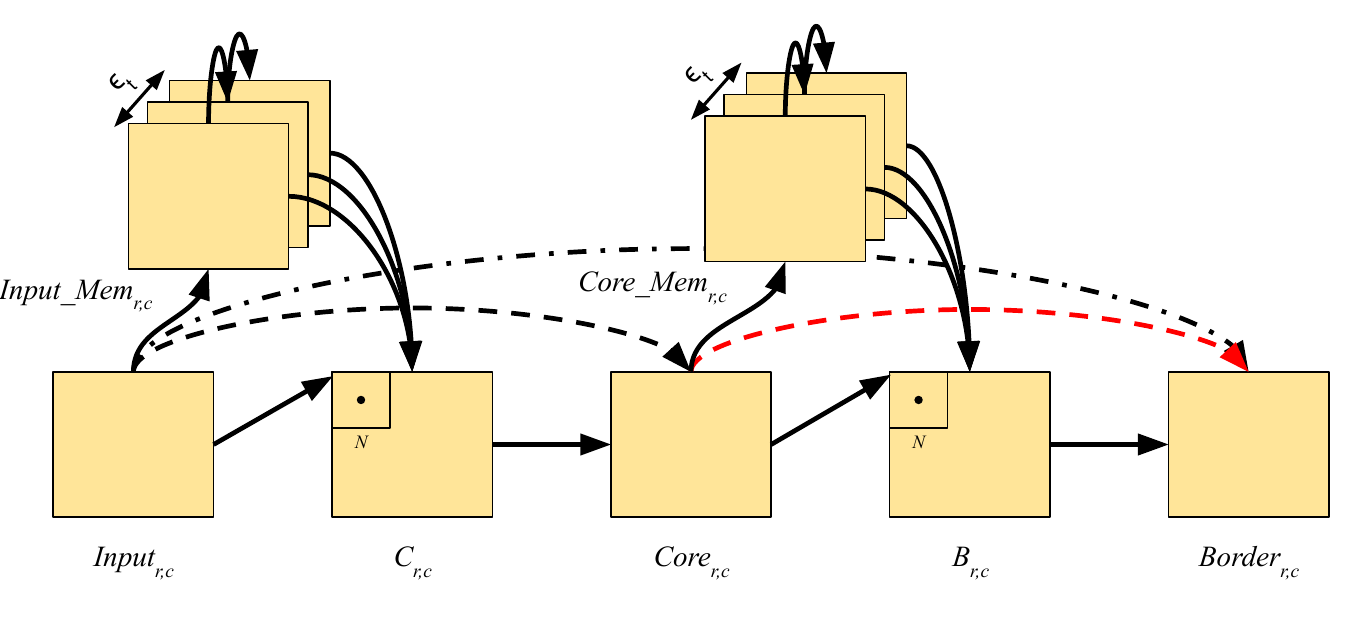}
\caption{Augmentations to Figure~\ref{fig:flat_connectivity} shown as $\epsilon_{t}$ sets of $R \times C$ $Input\_Mem$ and $R \times C$ $Core\_Mem$ neurons. The additional synaptic connectivity illustrates how prior frames of input and core classifications are persisted deeper into the network's runtime and are recurrently fed back into the compute portion of the network.}
\label{fig:temporal_flat_connectivity}
\end{center}
\end{figure*}

For both of the flat and systolic temporal algorithm augmentations, we introduce two new collections of neurons that are parameterized by $R$, $C$, and $\epsilon_{t}$: $Input\_Mem$ and $Core\_Mem$. The additional collections functionally act as memory and recurrence cells within the network. For the flat algorithm, there are $\epsilon_{t}$ $R \times C$ blocks of $Input\_Mem$ and $\epsilon_{t}$ $R \times C$ blocks of $Core\_Mem$ neurons whose purposes are to simply store and echo back into the compute portion of the network prior input frames and core classifications, respectively. Figure~\ref{fig:temporal_flat_connectivity} augments Figure~\ref{fig:flat_connectivity} to illustrate the new neuron collections and how they are connected within the network. The additional neuron collections and connectivity are paralleled in the systolic algorithm where there are $\epsilon_{t}$ $R(2\epsilon + 1)$ blocks of $Input\_Mem$ and $\epsilon_{t}$ $R(2\epsilon + 1)$ blocks of $Core\_Mem$ neurons that perform the same function. We omit a figure showing this for the systolic algorithm to save space, and instead, we note the functional similarities and point out the key differences. We show the total additional neurons required along with their thresholds -- all of which are 1 -- by the temporal flat and systolic algorithms in Table~\ref{tab:temporal_dbscan_neurons}.  We do the same for the synapses in Table~\ref{tab:temporal_dbscan_synapses}, showing their weights and delays. 

All of the additional synaptic weights are 1, and nearly all of the delays are as well. However, for the temporal systolic algorithm, there are four synaptic sets that have delay values of $C + 2\epsilon + 4$. These are from the $Input_{r,\epsilon}$ and $Core_{r,\epsilon}$ neurons to the $Input\_Mem_{0,r,\epsilon}$ and $Core\_Mem_{0,r,\epsilon}$ neurons, respectively. Additionally, we see the same delays for synapses connecting $Input\_Mem_{e_{t},r,\epsilon}$ neurons to $Input\_Mem_{e_{t} + 1,r,\epsilon}$ neurons -- the same is true for the $Core\_Mem$ neurons. The synapses effectively buffer previous frames of input or previous core classifications made in network memory by leveraging synaptic delay so that they may be recurrently fed back to the compute portion of the network at the proper time, assuming the processing of successive frames. It takes $C + 2\epsilon + 4$ timesteps to process one $R \times C$ frame of inputs with the systolic algorithm, so it follows that we desire the recurrent feedback into the network to occur exactly $C + 2\epsilon + 4$ timesteps later. 

\begin{table}[]
\caption{Temporally Augmented DBSCAN Neurons}
\label{tab:temporal_dbscan_neurons}
\begin{tabular}{l|cc}
\textbf{Neurons}            & \textbf{Count}                                      & \textbf{Thresholds} \\ \hline
\\ \textbf{Flat Algorithm}     &                                                     &                     \\ \cline{1-1}
Base Implementation         & $5RC$                                               & -                   \\
$Input\_Mem_{r,c}$          & $\epsilon_{t}RC$                                    & 1                   \\
$Core\_Mem_{r,c}$           & $\epsilon_{t}RC$                                    & 1                   \\
\textbf{Total}              & $RC(5 + 2\epsilon_{t})$                             &                     \\ \\ \hline
\\ \textbf{Systolic Algorithm} &                                                     &                     \\ \cline{1-1}
Base Implemenatation        & $R(4\epsilon + 5)$                                  & -                   \\
$Input\_Mem_{r,c}$          & $\epsilon_{t}R(2\epsilon + 1)$                      & 1                   \\
$Core\_Mem_{r,c}$           & $\epsilon_{t}R(2\epsilon + 1)$                      & 1                   \\
\textbf{Total}              & $R((4\epsilon + 5) + 2\epsilon_{t}(2\epsilon + 1))$ &                    
\end{tabular}
\end{table}

\begin{table*}[]
\centering
\caption{Temporally Augmented DBSCAN Synapses}
\label{tab:temporal_dbscan_synapses}
\begin{tabular}{lcrr}
\multicolumn{1}{l|}{\textbf{Synapse Sets}}                                                         & \textbf{Count}           & \textbf{Weight} & \textbf{Delay}      \\ \hline
\\ \textbf{Flat Algorithm}                                                                            &                          &                 &                     \\ \cline{1-1}
\multicolumn{1}{l|}{Base Implementation}                                                           & $RC(3 + 2N)$             & -               & -                   \\
\multicolumn{1}{l|}{$Input_{r,c} \rightarrow Input\_Mem_{0,r,c}$}                                  & $RC$                     & 1               & 1                   \\
\multicolumn{1}{l|}{$Input\_Mem_{e_{t},r,c} \rightarrow Input\_Mem_{e_{t}+1,r,c}$}                 & $RC(\epsilon_{t} - 1)$   & 1               & 1                   \\
\multicolumn{1}{l|}{$Input\_Mem_{e_{t},r,c} \rightarrow C_{r,c}$}                                 & $RC\epsilon_{t}N$        & 1               & 1                   \\
\multicolumn{1}{l|}{$Core_{r,c} \rightarrow Core\_Mem_{0,r,c}$}                                    & $RC$                     & 1               & 1                   \\
\multicolumn{1}{l|}{$Core\_Mem_{e_{t},r,c} \rightarrow Core\_Mem_{e_{t}+1,r,c}$}                   & $RC(\epsilon_{t} - 1)$   & 1               & 1                   \\
\multicolumn{1}{l|}{$Core\_Mem_{e_{t},r,c} \rightarrow B_{r,c}$}                                  & $RC\epsilon_{t}N$        & 1               & 1                   \\
\multicolumn{1}{l|}{\textbf{Total}}                                                                &    $RC(2(\epsilon_{t}+1)(N+1)+1)$                      &                 &                     \\ \hline
\\ \textbf{Systolic Algorithm}                                                                        &                          &                 &                     \\ \cline{1-1}
\multicolumn{1}{l|}{Base Implementation}                                                           & $R(2N + 4\epsilon + 3)$  & -               & -                   \\
\multicolumn{1}{l|}{$Input\_Mem_{e_{t},r,e} \rightarrow Input\_Mem_{e_{t},r,e-1}$}                 & $2\epsilon_{t}R\epsilon$ & 1               & 1                   \\
\multicolumn{1}{l|}{$Input_{r,\epsilon} \rightarrow Input\_Mem_{0,r,\epsilon}$}                    & $R$                      & 1               & $C + 2\epsilon + 4$ \\
\multicolumn{1}{l|}{$Input\_Mem_{e_{t},r,\epsilon} \rightarrow Input\_Mem_{e_{t} + 1,r,\epsilon}$} & $R(\epsilon_{t} - 1)$    & 1               & $C + 2\epsilon + 4$ \\
\multicolumn{1}{l|}{$Input\_Mem_{e_{t},i,e} \rightarrow C_{r}$}                                   & $\epsilon_{t}RN$         & 1               & 1                   \\
\multicolumn{1}{l|}{$Core\_Mem_{e_{t},r,e} \rightarrow Core\_Mem_{e_{t},r,e-1}$}                   & $2\epsilon_{t}R\epsilon$ & 1               & 1                   \\
\multicolumn{1}{l|}{$Core_{r,\epsilon} \rightarrow Core\_Mem_{0,r,\epsilon}$}                      & $R$                      & 1               & $C + 2\epsilon + 4$ \\
\multicolumn{1}{l|}{$Core\_Mem_{e_{t},r,\epsilon} \rightarrow Core\_Mem_{e_{t} + 1,r,\epsilon}$}   & $R(\epsilon_{t} - 1)$    & 1               & $C + 2\epsilon + 4$ \\
\multicolumn{1}{l|}{$Core\_Mem_{e_{t},i,e} \rightarrow B_{r}$}                                    & $\epsilon_{t}RN$         & 1               & 1                   \\ \hline
\multicolumn{1}{l|}{\textbf{Total}}         & $R(2\epsilon_t(2\epsilon + N + 1) + 4\epsilon + 2N + 3)$                       &                 & 
\end{tabular}
\end{table*}

\subsection{Optimizing Resource Usage}\label{sec:opt}
\noindent It is worth noting that the extra neuron collections are \textbf{only} necessary because the RISP neuroprocessor does not support multi-edges between any two neurons. With multi-edge support, the extra neuron hardware and much of the extra synaptic hardware can be omitted. For the temporal flat algorithm, the modification is trivial. Instead of the extra neurons and synapses shown in Tables~\ref{tab:temporal_dbscan_neurons} and~\ref{tab:temporal_dbscan_synapses}, only $2\epsilon_{t}RCN$ extra synapses are necessary. For the base implementation flat algorithm, there are $RC(N-1)$ synapses from the $Input_{r,c}$ neurons to the $C_{r,c}$ neurons and $RC(N-1)$ synapses from the $Core_{r,c}$ neurons to the $B_{r,c}$ neurons. With multi-edge support, we need only add $RCN$ synapses from the $Input_{r,c}$ neurons to the $C_{r,c}$ neurons and from the $Core_{r,c}$ neurons to the $B_{r,c}$ neurons $\epsilon_{t}$ times each. We use $N$ and not $N-1$ because older event activity occurring at $r,c$ \textbf{is} in the neighborhood~$N$ of event $E_{r,c}$. The delay values for each set of synapses starts at $2$ and increases up to $\epsilon_{t} + 1$ for each set of $2RCN$ synapses added to feed back information into the network. Previously, we needed unique neurons to capture and persist prior input frames and core classifications in their activation potentials as a sort of memory cell; however, with multi-edge support, we can simply add the extra synapses with increasing delay values to effect this memory. This generalization applies to the systolic implementation as well. However, its explanation is more complex, so we omit it for the sake of brevity and clarity, as the actual implementation of this optimization is beyond the scope of this work and limited by our choice of neuromorphic processor.

\section{Analysis}

\noindent In~\cite{rp:24:ani}, some examples of the full network sizes and the number of timesteps that it takes to fully process one input are displayed in Table 3. We do the same for our temporal network extensions in Table~\ref{tab:sizes}. We first present the general equations for neuronal and synaptic totals, runtime, and fan-in and fan-out. We then present the totals for two examples (similar to those presented in~\cite{rp:24:ani}):

\begin{enumerate}
    \item A $10 \times 10$ sized input with $\epsilon = \epsilon_{t} = 2$
    \item A $260 \times 346$ sized input with $\epsilon = \epsilon_{t} = 4$
\end{enumerate}

For the examples, we do not calculate the totals with the equations but instead construct the networks with our software and report the totals as such. Compared to the base implementations detailed in~\cite{rp:24:ani}, these networks are much larger, which is to be expected. The farther back in time we want to consider events as being in the neighborhood of $E_{r,c}$, the greater the number of $Input\_Mem$ and $Core\_Mem$ neurons and their requisite synaptic connectivity. This greatly motivates architectural optimization, such as that discussed in~\ref{sec:opt}.

\begin{table*}[ht]
\caption{\label{tab:sizes} Comparison of the spatiotemporal flat and systolic networks with example parameters. }
\begin{center}
{\small
%\rotatebox{90}{
\begin{tabular}{|l|ccc|}
\hline
 & & \textbf{Temporal Flat} & \\
 & General & $10 \times 10$ & $260 \times 346$ \\
\hline
Neurons & $RC(5 + 2\epsilon_{t})$ & 900 & 1,169,480  \\
Synapses & $RC(2(\epsilon_{t}+1)(N+1)+1)$ & 12,316 & 72,770,360 \\
Timesteps to Solution & 5 & 5 & 5 \\
Timesteps to Reuse & 1 & 1 & 1  \\
Max Synapse Delay & 4 & 4 & 4  \\
Max Neuron Threshold & $minPoints-1$ & $minPoints-1$ & $minPoints-1$ \\
Max Fan-Out & $N+2$ & 27 & 83 \\
Max Fan-In & $\epsilon_{t}(N) + N - 1$ & 74 & 404  \\
\hline
 & & \textbf{Temporal Systolic} & \\
 & General & $10 \times 10$ & $260 \times 346$  \\
\hline
Neurons & $R((4\epsilon + 5) + 2\epsilon_{t}(2\epsilon + 1))$ & 330 & 24,180 \\
Synapses & $R(2\epsilon_t(2\epsilon + N + 1) + 4\epsilon + 2N + 3)$ & 1,630 & 232,460 \\
Timesteps to Solution & $C+2\epsilon+4$ & 18 & 358 \\
Timesteps to Reuse & $C+2\epsilon$ & 14 & 354 \\
Max Synapse Delay & $C + 2\epsilon + 4$ & $C + 2\epsilon + 4$ & $C + 2\epsilon + 4$ \\
Max Neuron Threshold & $minPoints-1$ & $minPoints-1$ & $minPoints-1$ \\
Max Fan-Out & $N+3$ & 28 & 84 \\
Max Fan-In & $\epsilon_{t}(N) + N - 1$ & 74 & 404 \\
\hline

\end{tabular}}
%}
\end{center}
\end{table*}

\section{Partial Spatiotemporal DBSCAN}
\label{sec:partial}

\noindent With the goal of deploying these networks at the edge, we recognize that a network of 72 million synapses is likely prohibitive to embedded hardware. We discuss a partial, scalable implementation of the network in a method similar to that in~\cite{rp:24:ani}. In the same manner, we can perform the temporal DBSCAN operation on different sub-regions of the event camera's field-of-vision in order to create a smaller network, further trading time for space. From~\cite{rp:24:ani}, we know that this process is not as simple as partitioning the input into disjoint sub-inputs, and that in order to properly classify events that are on the borders of some region of interest, we need additional information from the $\epsilon$ border(s) or ring(s) surrounding the region of interest.

Furthermore, with the temporal DBSCAN extension, we want the prior $\epsilon_{t}$ frames to influence the classification of events in the current frame. Plainly, events in the top left corner of an input grid for the previous $\epsilon_{t}$ frames should contribute to the classifications of events in the top left corner of an input grid for the current frame. To accomplish this, we simply need to modify the delay values for the synapses responsible for persisting prior inputs (or frames) deeper into the network's runtime so that when processing the same region in a later frame, the current frame's inputs and core classifications are fed back into the compute portion of the network. Specifically, the synaptic delays that need to be adjusted are the following.

\begin{enumerate}
    \item $Input_{r,c} \rightarrow Input\_Mem_{0,r,c}$ 
    \item $Input\_Mem_{e_{t},r,c} \rightarrow Input\_Mem_{e_{t}+1,r,c}$ 
    \item $Core_{r,c} \rightarrow Core\_Mem_{0,r,c}$ 
    \item $Core\_Mem_{e_{t},r,c} \rightarrow Core\_Mem_{e_{t}+1,r,c}$ 
\end{enumerate}

Let's suppose we have an input frame whose size is $260 \times 346$ and that the size of our partial DBSCAN kernel is $I_{R} = 10, I_{C} = 10$. For a partial implementation of the flat, temporal DBSCAN SNN, it would take $ceil(\frac{R}{I_{R}}) * ceil(\frac{C}{I_{C}})$ timesteps to process an input frame sized $R\times C$. Therefore, assuming consistent algorithmic processing of subregions between frames (e.g. processing subregions in row-major order), setting the aforementioned delay values to $ceil(\frac{R}{I_{R}}) * ceil(\frac{C}{I_{C}}) = ceil(\frac{260}{10}) * ceil(\frac{346}{10}) = 910 $ will ensure that previous input events and core classifications of the same subregion some $e_{t}$ frames in the past will appropriately influence the classifications of the current input events in the same subregion in the current frame. It is similar for a partial implementation of the systolic, temporal DBSCAN SNN. It takes $ceil(\frac{R}{I_{R}}) * (C + 4\epsilon)$ timesteps to process an input frame sized $R\times C$; therefore, we set the delay values for the same synapse groups to $ceil(\frac{R}{I_{R}}) * (C + 4\epsilon)$. 

While it will take longer to process one input frame using the partial implementations of the temporal DBSCAN algorithm, the smaller network sizes overall may be desirable given a particular set of hardware resource constraints.

\section{Open-Source Implementation}

\noindent Our implementation of these spatiotemporal network constructions is available as open-source software at the same repository where our prior work from~\cite{rp:24:ani} is:~\url{https://github.com/TENNLab-UTK/dbscan}. 

\section{Conclusion}

\noindent We have presented spatiotemporal extensions for the ``flat'' and ``systolic'' neuromorphic DBSCAN algorithms that were presented in~\cite{rp:24:ani}. Furthermore, we have characterized their sizes and runtimes. We have additionally discussed an architectural advantage of multi-edge support between two neurons and presented partial implementations of the algorithmic extensions in order to target resource-constrained hardware.

Our future work involves comparing these spatiotemporal DBSCAN networks to our earlier spatial DBSCAN networks in terms of efficacy. We will also deploy multiple sizes of these parameterizable networks on available hardware to characterize their latency and throughput on data collected and processed in real-time from event cameras.

\bibliographystyle{plain}
\bibliography{bib}

\end{document}